\documentclass{article}

\usepackage{PRIMEarxiv}

\usepackage{placeins}
\usepackage[utf8]{inputenc} 
\usepackage[T1]{fontenc}    
\usepackage{hyperref}       
\usepackage{url}            
\usepackage{booktabs}       
\usepackage{amsfonts}       
\usepackage{nicefrac}       
\usepackage{microtype}      
\usepackage{lipsum}
\usepackage{fancyhdr}       
\usepackage{graphicx}       
\usepackage[round]{natbib}
\usepackage{array}
\usepackage{todonotes}
\graphicspath{{media/}}     
\usepackage{listingsutf8}
\usepackage{listing}
\usepackage{caption}
\usepackage{appendix}
\usepackage{chngcntr}
\usepackage{alphalph}
\usepackage{float}

\title{The HalluRAG Dataset: Detecting Closed-Domain Hallucinations in RAG Applications Using an LLM’s Internal States
}

\author{
  Fabian Ridder \\
  Computer Science Department\\
  University of Münster \\
  Münster, Germany\\
  \texttt{fridder@uni-muenster.de} \\
   \And
  Malte Schilling \\
  Computer Science Department\\
  University of Münster \\
  Münster, Germany\\
  \texttt{malte.schilling@uni-muenster.de} \\
}

\begin{document}

\maketitle


\begin{abstract}
Detecting hallucinations in large language models (LLMs) is critical for enhancing their reliability and trustworthiness. Most research focuses on hallucinations as deviations from information seen during training. However, the opaque nature of an LLM's parametric knowledge complicates the understanding of why generated texts appear ungrounded: The LLM might not have picked up the necessary knowledge from large and often inaccessible datasets, or the information might have been changed or contradicted during further training. Our focus is on hallucinations involving information not used in training, which we determine by using recency to ensure the information emerged after a cut-off date. This study investigates these hallucinations by detecting them at sentence level using different internal states of various LLMs. We present HalluRAG, a dataset designed to train classifiers on these hallucinations. Depending on the model and quantization, MLPs trained on HalluRAG detect hallucinations with test accuracies ranging up to $75\%$, with Mistral-7B-Instruct-v0.1 achieving the highest test accuracies. Our results show that IAVs detect hallucinations as effectively as CEVs and reveal that answerable and unanswerable prompts are encoded differently as separate classifiers for these categories improved accuracy. However, HalluRAG showed some limited generalizability, advocating for more diversity in datasets on hallucinations.
\end{abstract}

\textbf{Code} --- \href{https://github.com/F4biian/HalluRAG}{https://github.com/F4biian/HalluRAG} \\
\textbf{Dataset} ---\href{https://miami.uni-muenster.de/Record/10.17879\%2F84958668505/Details}{https://miami.uni-muenster.de/Record/10.17879\%2F84958668505/Details}


\section{Introduction}

Large language models (LLMs) are generative machine learning models that generate sequences of tokens one by one. As these are trained on massive datasets, LLMs have become excellent at producing coherent text. When sufficiently trained, these models can be prompted on novel tasks \citep{radford2019language}, meaning that one can apply an LLM on a different task---e.g., asking questions---to which the model generates a corresponding output as an answer. One impressive feature of LLMs is that the generated answers are not only coherent as a possible text but appear reasonable. While this allows LLMs as generative models to create novel outputs that were not part of the training data, there is a downside: The produced answers can be incorrect or not factual while being plausible-sounding answers. Another limitation of directly questioning LLMs is that they are restricted to the knowledge they learned from training data.

Retrieval-augmented generation (RAG)~\citep{lewis2021retrievalaugmented} is a commonly used technique aimed at overcoming these challenges \citep{shuster-etal-2021-retrieval-augmentation}. In RAG, background knowledge indexed beforehand is retrieved from a database as additional context and input to the model. For example, a user's question is enriched with particular snippets from external documents (e.g., PDF, PowerPoint, or text files) that serve as additional input to the LLM. RAG exploits the capability to use large contexts as inputs, integrating knowledge from the context into the answer. Typical examples of RAG systems are chatbots that provide customer support based on a database of past transactions \citep{xu2024rag} or tutoring systems in which context, as well as answers, are enhanced by infusing background knowledge from trusted sources \citep{levonian2023retrieval,kahl2024evaluating}. Although LLM applications incorporating RAG are aimed at hallucinating less, hallucinations still occur.
Generating different forms of non-factual answers---or those contradicting some of the available knowledge---remains a significant challenge for incorporating LLMs into workflows and services.

One approach to address hallucinations is to detect them, which allows the system to intervene. The system could refuse to return the hallucinated response and regenerate another response. 
In this work, we analyze an approach that leverages the LLM's internal states to determine if it currently fabricates hallucinations while answering a user's question. The analysis is based on a novel dataset for detecting closed-domain hallucinations---i.e., a dataset of queries that cover knowledge not included in training datasets but for which additional context can be provided as input to the LLM. Furthermore, we start by introducing a distinction between different types of hallucinations.

\begin{figure}[t!b]
    \centering
    \includegraphics[scale=0.36]{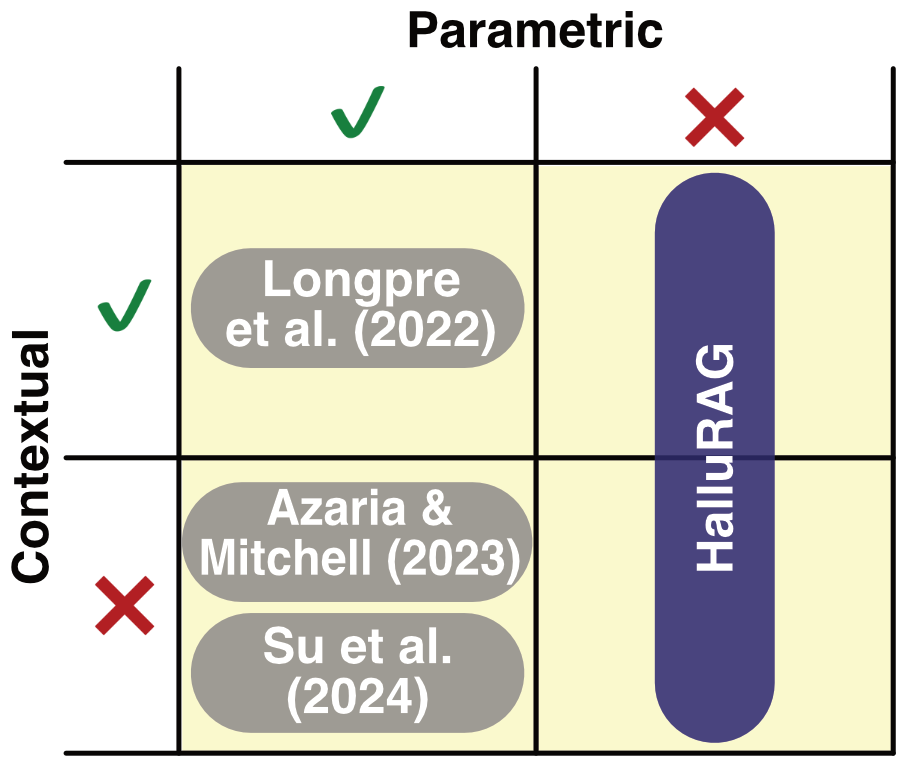}
    \caption{Differentiation of approaches~\citep{azaria2023internal, su2024unsupervised, longpre2022entitybased} based on the type of queried knowledge. Current methods in the literature focus on knowledge that is assumed as entrained into the LLM's parameters (parametric). This is often difficult to assess as, first, the training data is not always accessible. Second, it is not clear if and how this information was accurately learned by the model during training, for example, which level of detail was kept. Or how further training on other data might have influenced the information, for example, when contradicting pieces of information were present in the training data. In contrast, our focus (in the HalluRAG dataset) is on knowledge the LLM could not have seen during training, avoiding speculative assumptions. This gives us full control over offering this knowledge as context to the model or dealing with questions the model can not answer in any case. The second dimension distinguishes if relevant information for answering the question was provided as part of the context.}
    \label{fig:param-context}
\end{figure}

\section{Hallucination Types and Characteristics }
A hallucination is a statement that lacks grounding in the LLM's knowledge. Specifically, a statement is considered ungrounded if it does not exist in any form within the model's knowledge---neither as part of the training data nor as additional context (such as provided in RAG application). This makes an ungrounded statement ``fabricated''~\citep{agrawal2024language}. Notably, the determination of whether a statement constitutes a hallucination does not depend on its factual accuracy or alignment with current scientific consensus as such but rather on how well the response aligns with the knowledge sources that constitute the model's foundation.

In an LLM, there are two forms of knowledge and memories: First, during training, knowledge is integrated into the model's parameters from the training data. This is known as \textbf{parametric} knowledge. Second, contextual information can be provided to the model as part of the input or prompt, referred to as \textbf{contextual} knowledge~\citep{longpre2022entitybased}. 
When considering hallucinations, we should distinguish between these two kind of knowledge sources: parametric and contextual. On the one hand, absolute or \textbf{open-domain hallucinations} are generated statements with no grounding in an LLM's parametric knowledge or training data. This makes open-domain hallucinations difficult to detect due to the vast and often opaque nature of training datasets, which may be unpublished or lack detailed references. On the other hand, \textbf{closed-domain hallucinations} involve statements ungrounded with respect to the provided context and the knowledge given as input in a prompt~\citep{agrawal2024language, friel2023chainpollhighefficacymethod}. Additional distinctions for hallucinations have been introduced~\citep{zhang2023sirens} that address special cases~\citep{longpre2022entitybased} and lead to further subdivisions~\citep{maynez2020faithfulness}. From our point of view, it is important to explicitly distinguish how a hallucination fabricates new knowledge: Does it contradict or deviate from knowledge the model has seen during training and encoded in its parameters? This is the most widely assessed type of hallucination, where statements are actually false and contradict---or are assumed as contradicting---the training data (Fig.~\ref{fig:param-context}). Further research (see below on Related Work) has investigated hallucinations where LLMs fabricate statements that are false with respect to knowledge provided as additional input. HalluRAG is designed to specifically address knowledge that the model has not seen during training, allowing systematic control over whether information is provided as context.


\section{Related Work}
There are two main characteristics or dimensions of \textbf{hallucination detection methods}: first, the level of access to the LLM, and second, the use of references for comparison during the detection process.


Detection methods require different levels of access to an LLM. At one end of such a spectrum, \textbf{black-box} methods utilize only the LLM's output token, focusing exclusively on the generated language. On the other end, \textbf{white-box} methods tap into the model's internal states for detection, requiring full access to the LLM, which is possible when the model is run locally. \textbf{Gray-box} detection methods are an intermediate case, employing the output probability distribution for tokens---as provided as well by many online LLM services---but without requiring full access to internal states~\citep{manakul2023selfcheckgpt}.

The second characteristic differentiates whether references are available for comparison purposes. \textbf{Reference-based} (supervised) methods use a reference text to validate output accuracy, whereas \textbf{reference-free} (unsupervised or zero-reference) methods assess output without the need for any reference for comparison. This is particularly important, as in many cases, no reliable references are available~\citep{fang2024zeroresourcehallucinationdetectiontext}. White-box and gray-box approaches are typically applied without references, as hallucination detection should rely on the internal states or the output distribution alone, without considering the generated text itself.


Hallucination detection is a rapidly growing field of research, with an increasing number of proposed methods in all mentioned areas. Reference-based black-box methods were proposed early on. For example, RefChecker~\citep{hu-etal-2023-refchecker} evaluates `knowledge triplets' by comparing them to real references under various context conditions. Such methods have been further improved, e.g., \citep{agrawal2024language} used Bing to verify if suggested references by an LLM really exist. 
Reference-free black-box methods focus on the parametric knowledge of LLMs and prompt the LLM to either explicitly or implicitly evaluate generated answers by themselves:  \cite{kadavath2022language} proposed such self-evaluation to measure confidence in the models' answers. In contrast, SelfCheckGPT~\citep{manakul2023selfcheckgpt} checks for self-consistency across multiple generated outputs for the same prompt.
\cite{manakul2023selfcheckgpt} also suggested a reference-free gray-box method using likelihood and entropy metrics that allow to flag hallucinations based on the measured response randomness. 

\begin{figure}[t!b]
    \centering
    \includegraphics[scale=0.6]{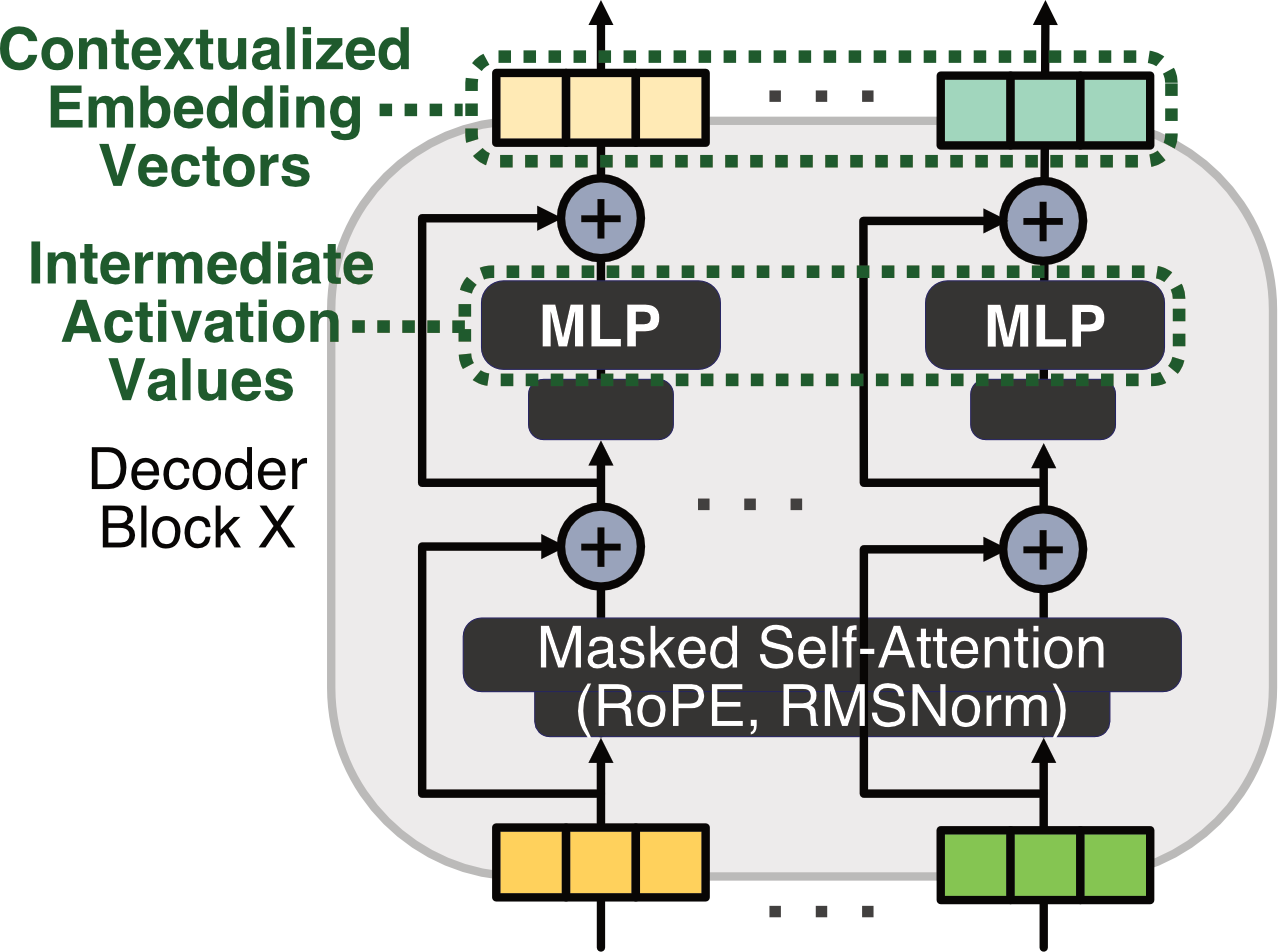}
    \caption{Locations of intermediate activation values and contextualized embedding vectors in the simplified architecture of LLaMA-2-7B including RMSNorm~\citep{zhang2019root} and Rotary Position Embeddings~\citep{su2023roformer}. While \cite{azaria2023internal} and \cite{su2024unsupervised} used contextualized embedding vectors as input to a binary classifier, we extend this approach by also considering intermediate activation values as classifier inputs.}
    \label{fig:llama-arch-classifier}
\end{figure}

Lastly, reference-free white-box methods involve analyzing the internal components of an LLM, such as the INSIDE EigenScore~\citep{chen2024inside} for assessing self-consistency through covariance analysis, the MIND classifier~\citep{su2024unsupervised}, and SAPLMA's template-based classifier~\citep{azaria2023internal}.
Closest to our approach is the work of \cite{su2024unsupervised} and \cite{azaria2023internal}, who follow a white-box, reference-free detection method. Both evaluate generated sentences solely by examining the model's internal states without requiring an external reference during production, but with full access to the LLM's internal structure. This is achieved by training a neural network that uses the internal states as an input and predicts the likelihood of a hallucination in the concurrently generated output sequences.
While our approach uses a similar setup, it specifically focuses on distinguishing closed-domain hallucinations and, as such, focuses only on knowledge queries that are outside of the training data (Fig.~\ref{fig:param-context}).
In contrast, \cite{su2024unsupervised} and \cite{azaria2023internal} assume that Wikipedia content is reliably encoded in an LLM's parametric knowledge, which seems reasonable for basic and foundational information, such as elements on the periodic table. However, this is not necessarily true for other types of knowledge, which might change over time or for which contradictory information has been encountered during training. Moreover, it is difficult to determine whether such knowledge---even if seen during training---has been accurately encoded in the LLM's parameters. Additionally, our proposed method takes into account intermediate activation values from the decoder blocks' multilayer perceptrons (MLPs), an internal state type that previous research has mostly neglected, as it primarily focused on contextualized embedding vectors (also referred to as activation values, see Fig.~\ref{fig:llama-arch-classifier}).

\section{Method}
We aim to train an MLP to identify sentence-level hallucinations by analyzing specific internal states in RAG applications. In particular, we focus on closed-domain hallucinations, which involve querying the model for answers that it cannot infer from its training. To train the classifier, we require a suitable supervised dataset that links the internal state---corresponding to a given input that lead to a generated sentence---to a hallucination label. 


While we aim for reference-free hallucination detection in RAG applications, \textbf{question-answering datasets} are suitable for training these methods. There are several fitting datasets, e.g., the NoMIRACL dataset~\citep{thakur2024nomiracl} and the RAGTruth dataset~\citep{wu2023ragtruth}. The latter offers $18,000$ word-level human-annotated hallucinations from LLM outputs across tasks such as summarization, question answering, and data-to-text, focusing on Mistral~\citep{jiang2023mistral} and LLaMA-2-models~\citep{touvron2023llama}. However, both datasets have some common drawbacks. First, the prompt formats are not diverse. Second, it is unclear whether information on questions and answers has been explicitly or potentially used during training of the LLM, as there are no criteria regarding recency.
Recency---considered as novel information that has emerged during recent times and was not available before a given cut-off date---is a crucial factor that is not guaranteed by either RAGTruth or NoMIRACL. Therefore, we created a novel dataset called HalluRAG, as we consider recency a promising means of determining what information an LLM may have been trained on. This enables us to focus exclusively on closed-domain hallucinations, explicitly controlling the answerability of prompts (Fig.~\ref{fig:param-context}). Since \cite{su2024unsupervised} found that the SAPLMA classifier by \cite{azaria2023internal} might have been overfitted, possibly due to its simplistic template-based approach, we aim for HalluRAG to encompass a broader variety of formats and topics. Our objective is to create a heterogeneous dataset that allows training MLP-based classifiers, which can be broadly applied to any RAG application. As \cite{su2024unsupervised} found a distinction between a forced hallucination and one that emerged naturally, hallucinations in HalluRAG will not be enforced but should only emerge, leading to \textbf{natural} hallucinations.

In the following, we will, first, present the HalluRAG dataset, which includes RAG prompts, generated responses, and internal states, fulfilling our requirements. Second, we will explain the training of an MLP-based classifier that uses the internal states of an LLM and estimates whether a generated sentence is hallucinated.

\begin{figure}[t!b]
    \centering
    \includegraphics[width=0.9\columnwidth]{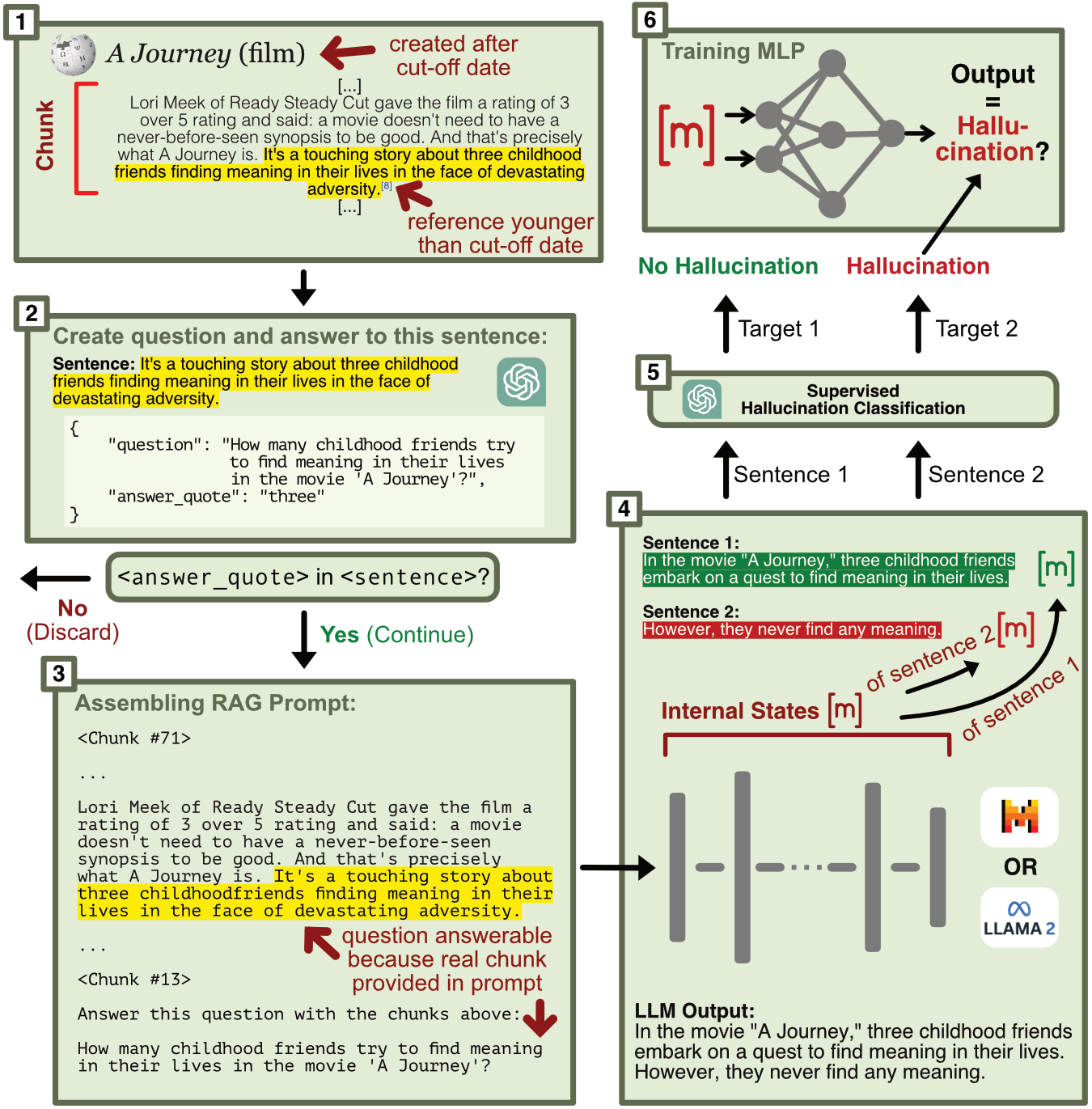}
    \caption{Overview of process flow for setting up the HalluRAG dataset: Shown is the whole process of a valid passage on Wikipedia turned into a RAG prompt, the corresponding generated sentences, and internal states for the HalluRAG dataset to eventually train a multilayer perceptron. For more details see text.}
    \label{fig:hallurag_creation}
\end{figure}

\subsection{Creating the HalluRAG Dataset}

We selected Wikipedia as a semi-structured source of information because of its wide variety of topics and the large number of references within its articles. Furthermore, Wikipedia provides detailed timestamps indicating when information was created or updated. This is particularly important, as we use recency to ensure that information could not have been used for training. We used Wikipedia timestamps to verify that information is recent and was not available---even in an older form or relying on similar older content---during training of an LLM based on the LLM's cut-off date. The process of generating HalluRAG (dataset is available at \href{https://doi.org/10.17879/84958668505}{https://doi.org/10.17879/84958668505}) is briefly illustrated in Figure~\ref{fig:hallurag_creation} and explained in detail below. 

\begin{enumerate}
    \item Extracting recent sentences from Wikipedia: For a given cut-off date---in our case  February 22, 2024---all newer articles from the English Wikipedia were considered. From all potential articles, sentences with at least one reference, that consisted of more than $50$ characters, and were not linked to other Wikipedia articles were collected. Furthermore, we checked each date, access date, and archive date in the references to ensure they were not older than the cut-off date. If all of these conditions were met, a sentence was considered a dataset candidate.
    
    \item Question generation: Given a dataset candidate and its surrounding text, we prompted GPT-4o (gpt-4o-2024-05-13)~\citep{openai2024gpt4} to generate a question as could have been raised in a RAG application. GPT-4o was further instructed to extract the corresponding answer from the dataset candidate's sentence, which was necessary for our hallucination annotation in training. Additionally, we use the extracted answer as an effective verification step, ensuring it was a substring of the original sentence to remove the chance of hallucinations at this stage. The prompt used for this process is shown in Listing~\ref{lst:prompt_template_wiki2qna} in the Appendix. 


    \item Creating pairs of questions and answers: For each question, we generated two RAG prompts---one \textbf{answerable} and one \textbf{unanswerable}---reflecting the real-world variability in retrieval systems, where the correct chunk may not always be returned---or the question itself may lack a definitive answer. The first prompt included the relevant passage from Wikipedia where the answer was present, while the second contained an unrelated chunk from another Wikipedia article. To add diversity, we randomly selected one of three prompt templates (shown in Listings~\ref{lst:prompt_template_langchain_hub}, \ref{lst:prompt_template_1}, and \ref{lst:prompt_template_2} in the Appendix), a chunk size (350, 550, or 750 characters), and the number of chunks per prompt (1, 3, or 5).

    \item Observing internal states: We passed these RAG prompts to an LLM to generate answers for each question, using a temperature of $0.0$ and a token limit of $500$. For HalluRAG, we used LLaMA-2-7B-Chat-HF~\citep{touvron2023llama} and Mistral-7B-Instruct-v0.1~\citep{jiang2023mistral}. LLaMA-2-13B-Chat-HF~\citep{touvron2023llama} was also tested, but due to observed extrinsic behavior, all of its results are shown only in the Appendix. During each generation, specific internal states (see next section) were extracted and stored for each sentence of the generated response: the contextualized embedding vectors from the last token’s middle and last decoder blocks (noted as `cev (middle)' and `cev (last)') and the intermediate activation values from the same blocks (denoted `iav (middle)' and `iav (last)').\footnote{We thank \cite{su2024unsupervised} for providing their code on GitHub and for their assistance.} Previous work has suggested that the final token embeddings best capture the essence of the preceding text, effectively compressing its content into a single vector~\citep{azaria2023internal, su2024unsupervised}. 
    
    \item Labeling of responses as hallucinations: To label each sentence as either hallucinated or not, we used GPT-4o (gpt-4o-2024-05-13) to compare each generated sentence against the entire response, the Wikipedia passage, and the quoted answer from step 2. GPT-4o was prompted with a detailed Chain-of-Thought (CoT) prompt~\citep{wei2023chainofthought} to set four booleans: \texttt{conflicting}, \texttt{grounded}, \texttt{has\_factual\_information}, and \texttt{no\_clear\_answer}. Each combination of these booleans was mapped to either $1$ (“hallucinated”), $0$ (“non-hallucinated”), or `None' (“invalid”) based on the prompt’s answerability (shown in Table~\ref{tab:hallu_mapping} as part of more detailed results and overview in the Appendix). Sentences labeled as `None' were withheld from training, testing, and validation. This four-boolean framework showed distinct advantages, as it made GPT-4o’s decision-making more transparent and forced it to break the task down. This helps mitigate GPT-4o-induced hallucinations, as the model justifies each boolean with verifiable substring checks. As a benchmark, we compared these with $274$ human-labeled sentences. GPT-4o achieved an F1-score of $96.05\%$ and an accuracy of $97.81\%$ (the six found `misclassifications' appeared as debatable cases to us). Overall, GPT-4o appears as a reliable evaluator for this task.
\end{enumerate}

In total, HalluRAG comprises $19,731$ validly annotated sentences generated by LLaMA-2-7B, LLaMA-2-13B, or Mistral-7B, along with their quantizations (float8, int8, and int4) (Appendix, Table~\ref{tab:auto_labeling_results}). Notably, LLaMA-2-13B was only run in quantized inference. As mentioned, we are only interested in natural hallucinations, not forced ones. Therefore, as a first observation, we report the hallucination rates for the different models: The LLaMA-2-7B configurations exhibit a stable hallucination rate of approximately $21\%$ on HalluRAG, whereas Mistral-7B demonstrates a significantly lower hallucination rate of about $10\%$ (for full details see Table~\ref{tab:hallu_rates}, provided in the Appendix). When grouping by prompt template, we observed that the hallucination rate varied considerably. For instance, LLaMA-2-7B showed hallucination rates between $32\%$ and $40\%$ with a template from the Langchain hub (see Appendix, Listing~\ref{lst:prompt_template_langchain_hub}), compared to around $16\%$ with template 1 (see Appendix, Listing~\ref{lst:prompt_template_1}). This indicates that prompt engineering is already a straightforward yet effective method for reducing hallucinations.


\subsection{Training a Classifier for Hallucination Detection}
We trained a neural network as a classifier on the HalluRAG dataset (code for training and setting up dataset available at \href{https://github.com/F4biian/HalluRAG}{https://github.com/F4biian/HalluRAG}). The input to the network consists of different internal states from an LLM, obtained during the generation of our dataset. The target for the classifier is the assigned label, indicating whether the answer is hallucinated or non-hallucinated. The MLP structure consists of four linear layers (\texttt{input\_size}---256---128---64---1) with ReLU activation and a final sigmoid function, following
the structure used in MIND~\citep{su2024unsupervised} and SAPLMA~\citep{azaria2023internal}. The learning rate was set to $2.5\mathrm{e}{-6}$, with a weight decay of $1\mathrm{e}{-5}$ and an initial dropout rate of $15\%$ (similar to MIND's $20\%$). We employed early-stopping: training stopped after a maximum of $800$ epochs or when the validation loss did not improve for $30$ epochs. We proceeded with the checkpoint that had the lowest validation loss for testing. Ten MLPs were trained per LLM configuration to account for statistical variability.
We analyzed the mean and standard deviation of test accuracies across all these models.

However, the number of generated sentences by the LLMs varied across our different configurations (chunk size, chunk count, template, and answerability), inferring an imbalance in the data. To address this, we oversampled based on the configuration parameters and the target (hallucinated vs. non-hallucinated). In this way, we ensured a balance of one-to-one with respect to answerability and hallucination across training, validation, and test sets. For chunk sizes, numbers of chunks per prompt, and prompt templates, we ensured a ratio of one-third in each case. This approach enables interpretable and comparable test accuracies. 


\begin{table*}[t!]
\centering
\begin{tabular}{|l||r|r|r|r|r|}

\hline
\textbf{Internal State} & \textbf{All (\%)} & \textbf{None (\%)} & \textbf{float8 (\%)} & \textbf{int8 (\%)} & \textbf{int4 (\%)} \\
\hline

\textbf{LLaMA-2-7B-Chat-HF} & & & & & \\
cev (middle) & 65.41±0.87 & 62.82±0.55 & 63.57±1.40 & 62.79±1.82 & 71.01±0.59 \\
cev (last) & 60.40±1.79 & 60.93±1.05 & 57.61±2.03 & 59.68±4.08 & 71.24±2.29 \\
iav (middle) & 65.98±1.66 & 64.62±1.43 & 61.40±3.59 & 62.40±2.03 & 71.46±0.67 \\
iav (last) & 64.93±2.34 & 62.60±1.67 & 58.57±2.52 & 62.55±1.38 & 71.07±0.94 \\
\hline

\textbf{Mistral-7B-Instruct-v0.1} & & & & & \\
cev (middle) & 67.47±1.16 & \textit{66.98±4.29} & \textit{54.29±7.85} & \textit{65.26±0.62} & \textit{68.59±5.26} \\
cev (last) & 73.28±3.51 & \textit{67.27±4.83} & \textit{67.28±9.13} & \textit{68.02±7.70} & \textit{75.16±1.84} \\
iav (middle) & 69.95±2.38 & \textit{56.33±1.66} & \textit{61.35±2.75} & \textit{67.31±0.36} & \textit{71.94±5.97} \\
iav (last) & 74.91±0.92 & \textit{65.84±3.62} & \textit{70.58±5.03} & \textit{71.43±3.00} & \textit{78.94±1.88} \\
\hline
\end{tabular}
\caption{Average test accuracies and standard deviations of training ten MLPs on the internal states of the responses taken from HalluRAG. The column names indicate the quantization, where `None' means the use of no quantization and `All' considers all quantizations together (None, float8, int8, and int4). All internal states have been extracted from the last token's middle and last decoder block or `layer.' We abbreviate contextualized embedding vectors as `CEV' and intermediate activation values as `IAV'. The values based on a single quantized Mistral configuration are italic, since their test sets appear insufficiently small for being analyzed in more detail. See Table~\ref{tab:hallurag-l13} for LLaMA-2-13B-Chat-HF and Table~\ref{tab:hallurag-combinations} for combined internal states in the Appendix}
\label{tab:hallurag}
\end{table*}

\section{Results}
\subsection{Classifier Results on HalluRAG}
For each configuration (different LLMs, quantizations), ten classifiers were trained independently to determine whether a given input---provided as the internal states of the LLM---should be deemed as a hallucination. The neural networks were trained on balanced datasets, using oversampling as described above. An overview of the results is provided in table~\ref{tab:hallurag}. For the LLaMA-2-7B-Chat-HF model, test accuracies are around or above $60\%$, with int4 quantizations achieving a consistently higher accuracy of $71\%$. Notably, some larger-than-expected standard deviations make comparing different layers and quantizations challenging, but emphasize the importance of multiple training runs.

When comparing different layers, the middle layers of both, contextualized embedding vectors (CEVs) and intermediate activation values (IAVs), generally show slightly higher average accuracies than the last layers. 

In contrast, the Mistral-7B-Instruct-v0.1 model showed a different pattern compared to LLaMA-2-7B: the last layer's CEVs and IAVs achieved accuracies above $70\%$, while the middle layers remained below $70\%$. Importantly, in nearly all experiments, MLPs trained on sentences generated by Mistral-7B outperformed those trained on LLaMA-2-7B data.
Test accuracies for LLaMA-2-13B-Chat-HF yielded results at chance level of around 50\%~(see Appendix, table~\ref{tab:hallurag-l13}). Overall, LLaMA-2-13B-Chat-HF exhibited erratic behavior throughout the project by revealing poor hallucination detection accuracies (see Table\ref{tab:hallurag-l13}), contrary to other research findings~\citep{su2024unsupervised}. 

In particular, regardless of the specific model, all int4 quantizations consistently demonstrate the highest test accuracies in Table~\ref{tab:hallurag}, as well as in the other results~(given in the Appendix, tables~\ref{tab:hallurag-l13}, \ref{tab:hallurag-combinations}, \ref{tab:hallurag-answerable-all}, and \ref{tab:hallurag-unanswerable-all}). Furthermore, in combining internal states by concatenating their vectors does not seem to boost test accuracies considerably (Table~\ref{tab:hallurag-combinations}, Appendix).

We also tested the impact of withholding specific parameters (e.g., a chunk size of $350$ or prompt template 2) by excluding them from training and validation, and then evaluating on the full test set, including the previously withheld parameters. The results (given in the Appendix, table~\ref{tab:test_accuracies_left_out}) show no significant impact on test accuracy, except when training and validating solely on answerable questions, which lead to accuracies no better than random guessing.

\begin{table}[t!]
\centering
\begin{tabular}{|l||r|r|}
\hline
\textbf{Internal State} & \textbf{H-H-R (\%)} & \textbf{R-R-H (\%)} \\
\hline

\textbf{LLaMA-2-7B-Chat-HF} & & \\ 
cev (middle) & 51.04±0.30 & 52.84±1.34 \\
cev (last) & 49.58±0.51 & 50.24±0.59 \\
iav (middle) & 52.66±0.42 & 54.60±0.93 \\
iav (last) & 49.70±0.14 & 51.16±1.18 \\
\hline

\textbf{Mistral-7B-Instruct-v0.1} & & \\
cev (middle) & 56.60±0.79 & 62.27±0.97 \\
cev (last) & 57.11±1.56 & 58.46±0.33 \\
iav (middle) & 55.08±1.70 & 64.01±0.23 \\
iav (last) & 55.88±0.25 & 58.65±0.02 \\
\hline
\end{tabular}
\caption{Testing for generalization on a different dataset: Average test accuracies and standard deviations of testing ten MLPs, which on the one side have been trained and validated on HalluRAG but tested on RAGTruth (H-H-R), and on the other side trained and validated on RAGTruth but tested on HalluRAG (R-R-H). All quantizations together were used. All internal states have been extracted from the last token's middle and last decoder block or `layer.' We abbreviate contextualized embedding vectors as `CEV' and intermediate activation values as `IAV' (see Appendix, table~\ref{tab:tested-opposite-l13} for results on LLaMA-2-13B-Chat-HF.} 
\label{tab:tested-opposite}
\end{table}

\subsection{Generalization in Hallucination Classifiers}
We further cross-tested the MLPs trained on HalluRAG against RAGTruth~\citep{wu2023ragtruth}, as a similar dataset that however lacks a clear answerability distinction, and vice versa. This generally showed a poor performance for the LLaMA-2-7B model, which performed close to chance level (Table~\ref{tab:tested-opposite}). Mistral-7B again showed much better performance when compared to the LLaMA-2-7B model, but still performed worse than when trained on its specific dataset. This might indicate a general problem of overfitting in both HalluRAG- and RAGTruth-trained MLPs, with Mistral-7B demonstrating at least better-than-random test accuracies.


\begin{table}[t!]
\centering
\begin{tabular}{|l||r|r|}
\hline
\textbf{Internal State} & \textbf{Answ. (\%)} & \textbf{Unansw. (\%)}  \\
\hline

\textbf{LLaMA-2-7B-Chat-HF} & & \\
cev (middle) & 75.21±0.45 & 82.89±0.43 \\
cev (last) & 77.64±0.63 & 87.13±0.40 \\
iav (middle) & 75.52±0.59 & 81.98±0.83 \\
iav (last) & 78.22±0.54 & 86.26±0.14 \\
\hline

\textbf{Mistral-7B-Instruct-v0.1} & & \\
cev (middle) & 69.56±0.16 & 99.55±0.24 \\
cev (last) & 70.66±0.53 & 100.00±0.00 \\
iav (middle) & 68.55±2.08 & 99.87±0.04 \\
iav (last) & 70.38±0.33 & 100.00±0.00 \\
\hline
\end{tabular}
\caption{Average test accuracies and standard deviations of training ten MLPs on internal states of the responses based on either answerable or unanswerable questions of all LLM configurations (None, float8, int8, and int4). All internal states have been extracted from the last token's middle and last decoder block or `layer.' We abbreviate contextualized embedding vectors as `CEV' and intermediate activation values as `IAV'. Detailed results are provided in the Appendix, table~\ref{tab:hallurag-answerable-all} and Table~\ref{tab:hallurag-unanswerable-all}.}
\label{tab:hallurag-un-and-answerable}
\end{table}

\subsection{Training Separate Classifiers for Answerable and Unanswerable Questions}

The HalluRAG dataset distinguishes answerable and unanswerable questions. We trained, validated and tested classifiers separately on both partitions of the balanced dataset. The results (Table~\ref{tab:hallurag-un-and-answerable}) demonstrate a significant boost in test accuracies when using separate classifiers. For answerable questions, accuracies improved notably compared to the earlier results (Table~\ref{tab:hallurag}). For example, LLaMA-2-7B now achieves test accuracies above $75\%$. However, as an exemption, for Mistral-7B, the accuracy of the last layer drops from around $74\%$ to $70\%$. For unanswerable questions, the improvement is even more pronounced, with LLaMA-2-7B reaching over $80\%$ accuracy, and int8 as well as int4 models exceeding $90\%$ (Table~\ref{tab:hallurag-unanswerable-all}, Appendix). 

While training MLPs separately might simplify the problem, the classification remains non-trivial, particularly for answerable questions. For unanswerable questions, however, MLPs primarily need to determine from the internal states whether the LLM has responded with an ``I don't know'' (non-hallucinated) or something else (hallucinated). 
For instance, the sentence \textit{`I apologize, but I cannot provide an answer to your question as it is not based on any factual information provided in the context.'} by LLaMA-2-7B as a response to an unanswerable question is considered non-hallucinated. Same applies to its grounded response \textit{`The statue of Queen Elizabeth II in Oakham was funded through public subscription by Rutland people.'} to an answerable question.
Notably, Mistral-7B MLPs achieved near-perfect classification, with accuracies approaching $100\%$. While this shifts the problem to detecting prompt answerability, it still offers valuable insights into LLM behavior, as discussed later.

We further found that passing multiple internal states to our MLPs simultaneously did not improve performance in either case (Appendix, tables~\ref{tab:hallurag-answerable-all}~and~\ref{tab:hallurag-unanswerable-all}).


\section{Discussion and Conclusion}
In this work, we introduced HalluRAG, a dataset designed to detect closed-domain hallucinations in RAG systems by employing recency. We are focusing on novel information that were added after a specified cut-off data, and could not have been used for training. HalluRAG leverages an auto-labeling process using GPT-4o to annotate hallucinated and non-hallucinated sentences, achieving an F1-score of $96.05\%$. 
Even though HalluRAG aimed for diverse topics and formats, our evaluation demonstrates that the size of HalluRAG and the usage of $1,080$ prompts per LLM configuration appears still too small and the missing diversity still limit the MLPs' generalizability. Furthermore, GPT-4o-generated questions, based on Wikipedia passages, closely mirror the wording of the answers in these passages, which differs from real-world scenarios. 

Training a classifier for hallucinations on HalluRAG demonstrated moderate success, with classification accuracies significantly above chance level, ranging from $60\%$ to $75\%$.
Even though these results appear not sufficient on their own as a safeguard for hallucination detection, they clearly demonstrate that LLM layers contain information about the likelihood of hallucinations. While this behavior had been established for CEVs, it is a new finding for IAVs, further indicating that this information is broadly distributed inside of an LLM. Notably, adding more internal states to the MLP input did not substantially enhance performance, suggesting that a single internal state captures as much relevant information as multiple internal states in most cases. Further research could investigate the performance across additional layers.


When testing for generalization capabilities of the trained classifiers on the RAGTruth dataset, these performed poorly. Similarly, MLPs trained on the RAGTruth dataset did not generalize well when tested on HalluRAG. This suggests that there might be overfitting for both MLP types (Table~\ref{tab:tested-opposite}). From our point of view, this further emphasizes the need to create a wide variety of training data for hallucinations, using diverse prompt templates with substantial differences in format and length, rather than minor format changes. It would also be interesting to investigate the results of training and testing on a merged dataset combining RAGTruth and HalluRAG.

A surprising finding is that hallucinations in answerable and unanswerable prompts appear to be encoded differently, as test accuracies considerably increase for both cases when trained and tested separately (Table~\ref{tab:hallurag-un-and-answerable}). 
Training separately appears to enable the classifier to focus specifically on one type of hallucination, thereby boosting accuracy. This distinction should be further investigated to gain a better understanding of how these types differ. 


Importantly, test accuracies for Mistral-7B-Instruct-v0.1 are consistently higher than those for the LLaMA-2 series, indicating that the trained MLPs detect hallucinations more easily from Mistral-7B's internal states. This suggests that the likelihood of a hallucination is more clearly encoded in the internal states of Mistral-7B. We hypothesize that the training approach of Mistral-7B contributed to this in two ways. First, the distinct training approach prioritizes the efficient and thorough use of its 7 billion parameters rather than simply increasing the parameter count~\citep{jiang2023mistral}. Second, a dataset of higher quality for improved responses was used. This might have resulted in a clearer and more effective representation of language within Mistral-7B, making it easier for an MLP to distinguish between a hallucination and a non-hallucination. 

\section{Ethical Statement}
We constructed the HalluRAG dataset using publicly available Wikipedia articles, which, while open-source and potentially prone to inaccuracies or biases, ensure no reliance on personal data. Auto-labeling for hallucinated and non-hallucinated sentences was performed using GPT-4o, whose inherent biases may affect the dataset's fairness. To promote transparency, all data and code are publicly available for review and responsible use.

\bibliography{ridder2024hallurag}

\begin{thebibliography}{26}
\providecommand{\natexlab}[1]{#1}
\providecommand{\url}[1]{\texttt{#1}}
\expandafter\ifx\csname urlstyle\endcsname\relax
  \providecommand{\doi}[1]{doi: #1}\else
  \providecommand{\doi}{doi: \begingroup \urlstyle{rm}\Url}\fi

\bibitem[Agrawal et~al.(2024)Agrawal, Suzgun, Mackey, and
  Kalai]{agrawal2024language}
Ayush Agrawal, Mirac Suzgun, Lester Mackey, and Adam~Tauman Kalai.
\newblock Do language models know when they're hallucinating references?, 2024.

\bibitem[Azaria and Mitchell(2023)]{azaria2023internal}
Amos Azaria and Tom Mitchell.
\newblock The internal state of an llm knows when it's lying, 2023.

\bibitem[Chen et~al.(2024)Chen, Liu, Chen, Gu, Wu, Tao, Fu, and
  Ye]{chen2024inside}
Chao Chen, Kai Liu, Ze~Chen, Yi~Gu, Yue Wu, Mingyuan Tao, Zhihang Fu, and
  Jieping Ye.
\newblock Inside: Llms' internal states retain the power of hallucination
  detection, 2024.

\bibitem[Fang et~al.(2024)Fang, Huang, Tian, Fang, Pan, Fang, Wen, Pan, and
  Li]{fang2024zeroresourcehallucinationdetectiontext}
Xinyue Fang, Zhen Huang, Zhiliang Tian, Minghui Fang, Ziyi Pan, Quntian Fang,
  Zhihua Wen, Hengyue Pan, and Dongsheng Li.
\newblock Zero-resource hallucination detection for text generation via
  graph-based contextual knowledge triples modeling, 2024.
\newblock URL \url{https://arxiv.org/abs/2409.11283}.

\bibitem[Friel and Sanyal(2023)]{friel2023chainpollhighefficacymethod}
Robert Friel and Atindriyo Sanyal.
\newblock Chainpoll: A high efficacy method for llm hallucination detection,
  2023.
\newblock URL \url{https://arxiv.org/abs/2310.18344}.

\bibitem[Hu et~al.(2024)Hu, Ru, Qiu, Guo, Zhang, Xu, Luo, Liu, Zhang, and
  Zhang]{hu-etal-2023-refchecker}
Xiangkun Hu, Dongyu Ru, Lin Qiu, Qipeng Guo, Tianhang Zhang, Yang Xu, Yun Luo,
  Pengfei Liu, Yue Zhang, and Zheng Zhang.
\newblock Refchecker: Reference-based fine-grained hallucination checker and
  benchmark for large language models.
\newblock \emph{arXiv preprint arXiv:2405.14486}, 2024.

\bibitem[Jiang et~al.(2023)Jiang, Sablayrolles, Mensch, Bamford, Chaplot,
  de~las Casas, Bressand, Lengyel, Lample, Saulnier, Lavaud, Lachaux, Stock,
  Scao, Lavril, Wang, Lacroix, and Sayed]{jiang2023mistral}
Albert~Q. Jiang, Alexandre Sablayrolles, Arthur Mensch, Chris Bamford,
  Devendra~Singh Chaplot, Diego de~las Casas, Florian Bressand, Gianna Lengyel,
  Guillaume Lample, Lucile Saulnier, L{\'e}lio~Renard Lavaud, Marie-Anne
  Lachaux, Pierre Stock, Teven~Le Scao, Thibaut Lavril, Thomas Wang,
  Timoth{\'e}e Lacroix, and William~El Sayed.
\newblock Mistral 7b, 2023.

\bibitem[Kadavath et~al.(2022)Kadavath, Conerly, Askell, Henighan, Drain,
  Perez, Schiefer, Hatfield-Dodds, DasSarma, Tran-Johnson, Johnston, El-Showk,
  Jones, Elhage, Hume, Chen, Bai, Bowman, Fort, Ganguli, Hernandez, Jacobson,
  Kernion, Kravec, Lovitt, Ndousse, Olsson, Ringer, Amodei, Brown, Clark,
  Joseph, Mann, McCandlish, Olah, and Kaplan]{kadavath2022language}
Saurav Kadavath, Tom Conerly, Amanda Askell, Tom Henighan, Dawn Drain, Ethan
  Perez, Nicholas Schiefer, Zac Hatfield-Dodds, Nova DasSarma, Eli
  Tran-Johnson, Scott Johnston, Sheer El-Showk, Andy Jones, Nelson Elhage,
  Tristan Hume, Anna Chen, Yuntao Bai, Sam Bowman, Stanislav Fort, Deep
  Ganguli, Danny Hernandez, Josh Jacobson, Jackson Kernion, Shauna Kravec,
  Liane Lovitt, Kamal Ndousse, Catherine Olsson, Sam Ringer, Dario Amodei, Tom
  Brown, Jack Clark, Nicholas Joseph, Ben Mann, Sam McCandlish, Chris Olah, and
  Jared Kaplan.
\newblock Language models (mostly) know what they know, 2022.

\bibitem[Kahl et~al.(2024)Kahl, L{\"o}ffler, Maciol, Ridder, Schmitz, Spanagel,
  Wienkamp, Burgahn, and Schilling]{kahl2024evaluating}
Sebastian Kahl, Felix L{\"o}ffler, Martin Maciol, Fabian Ridder, Marius
  Schmitz, Jennifer Spanagel, Jens Wienkamp, Christopher Burgahn, and Malte
  Schilling.
\newblock Evaluating the impact of advanced llm techniques on ai-lecture tutors
  for a robotics course.
\newblock \emph{arXiv preprint arXiv:2408.04645}, 2024.

\bibitem[Levonian et~al.(2023)Levonian, Li, Zhu, Gade, Henkel, Postle, and
  Xing]{levonian2023retrieval}
Zachary Levonian, Chenglu Li, Wangda Zhu, Anoushka Gade, Owen Henkel,
  Millie-Ellen Postle, and Wanli Xing.
\newblock Retrieval-augmented generation to improve math question-answering:
  Trade-offs between groundedness and human preference.
\newblock \emph{arXiv preprint arXiv:2310.03184}, 2023.

\bibitem[Lewis et~al.(2021)Lewis, Perez, Piktus, Petroni, Karpukhin, Goyal,
  K{\"u}ttler, Lewis, tau Yih, Rockt{\"a}schel, Riedel, and
  Kiela]{lewis2021retrievalaugmented}
Patrick Lewis, Ethan Perez, Aleksandra Piktus, Fabio Petroni, Vladimir
  Karpukhin, Naman Goyal, Heinrich K{\"u}ttler, Mike Lewis, Wen tau Yih, Tim
  Rockt{\"a}schel, Sebastian Riedel, and Douwe Kiela.
\newblock Retrieval-augmented generation for knowledge-intensive nlp tasks,
  2021.

\bibitem[Longpre et~al.(2022)Longpre, Perisetla, Chen, Ramesh, DuBois, and
  Singh]{longpre2022entitybased}
Shayne Longpre, Kartik Perisetla, Anthony Chen, Nikhil Ramesh, Chris DuBois,
  and Sameer Singh.
\newblock Entity-based knowledge conflicts in question answering, 2022.

\bibitem[Manakul et~al.(2023)Manakul, Liusie, and
  Gales]{manakul2023selfcheckgpt}
Potsawee Manakul, Adian Liusie, and Mark J.~F. Gales.
\newblock Selfcheckgpt: Zero-resource black-box hallucination detection for
  generative large language models, 2023.

\bibitem[Maynez et~al.(2020)Maynez, Narayan, Bohnet, and
  McDonald]{maynez2020faithfulness}
Joshua Maynez, Shashi Narayan, Bernd Bohnet, and Ryan McDonald.
\newblock On faithfulness and factuality in abstractive summarization, 2020.

\bibitem[OpenAI et~al.(2024)OpenAI, Achiam, Adler, Agarwal, Ahmad, Akkaya,
  Aleman, Almeida, Altenschmidt, Altman, Anadkat, Avila, Babuschkin, Balaji,
  Balcom, Baltescu, Bao, Bavarian, Belgum, Bello, Berdine, Bernadett-Shapiro,
  Berner, Bogdonoff, Boiko, Boyd, Brakman, Brockman, Brooks, Brundage, Button,
  Cai, Campbell, Cann, Carey, Carlson, Carmichael, Chan, Chang, Chantzis, Chen,
  Chen, Chen, Chen, Chen, Chess, Cho, Chu, Chung, Cummings, Currier, Dai,
  Decareaux, Degry, Deutsch, Deville, Dhar, Dohan, Dowling, Dunning, Ecoffet,
  Eleti, Eloundou, Farhi, Fedus, Felix, Fishman, Forte, Fulford, Gao, Georges,
  Gibson, Goel, Gogineni, Goh, Gontijo-Lopes, Gordon, Grafstein, Gray, Greene,
  Gross, Gu, Guo, Hallacy, Han, Harris, He, Heaton, Heidecke, Hesse, Hickey,
  Hickey, Hoeschele, Houghton, Hsu, Hu, Hu, Huizinga, Jain, Jain, Jang, Jiang,
  Jiang, Jin, Jin, Jomoto, Jonn, Jun, Kaftan, Kaiser, Kamali, Kanitscheider,
  Keskar, Khan, Kilpatrick, Kim, Kim, Kim, Kirchner, Kiros, Knight, Kokotajlo,
  Kondraciuk, Kondrich, Konstantinidis, Kosic, Krueger, Kuo, Lampe, Lan, Lee,
  Leike, Leung, Levy, Li, Lim, Lin, Lin, Litwin, Lopez, Lowe, Lue, Makanju,
  Malfacini, Manning, Markov, Markovski, Martin, Mayer, Mayne, McGrew,
  McKinney, McLeavey, McMillan, McNeil, Medina, Mehta, Menick, Metz,
  Mishchenko, Mishkin, Monaco, Morikawa, Mossing, Mu, Murati, Murk, M{\'e}ly,
  Nair, Nakano, Nayak, Neelakantan, Ngo, Noh, Ouyang, O'Keefe, Pachocki, Paino,
  Palermo, Pantuliano, Parascandolo, Parish, Parparita, Passos, Pavlov, Peng,
  Perelman, de~Avila Belbute~Peres, Petrov, de~Oliveira~Pinto, Michael,
  Pokorny, Pokrass, Pong, Powell, Power, Power, Proehl, Puri, Radford, Rae,
  Ramesh, Raymond, Real, Rimbach, Ross, Rotsted, Roussez, Ryder, Saltarelli,
  Sanders, Santurkar, Sastry, Schmidt, Schnurr, Schulman, Selsam, Sheppard,
  Sherbakov, Shieh, Shoker, Shyam, Sidor, Sigler, Simens, Sitkin, Slama, Sohl,
  Sokolowsky, Song, Staudacher, Such, Summers, Sutskever, Tang, Tezak,
  Thompson, Tillet, Tootoonchian, Tseng, Tuggle, Turley, Tworek, Uribe,
  Vallone, Vijayvergiya, Voss, Wainwright, Wang, Wang, Wang, Ward, Wei,
  Weinmann, Welihinda, Welinder, Weng, Weng, Wiethoff, Willner, Winter,
  Wolrich, Wong, Workman, Wu, Wu, Wu, Xiao, Xu, Yoo, Yu, Yuan, Zaremba,
  Zellers, Zhang, Zhang, Zhao, Zheng, Zhuang, Zhuk, and Zoph]{openai2024gpt4}
OpenAI, Josh Achiam, Steven Adler, Sandhini Agarwal, Lama Ahmad, Ilge Akkaya,
  Florencia~Leoni Aleman, Diogo Almeida, Janko Altenschmidt, Sam Altman,
  Shyamal Anadkat, Red Avila, Igor Babuschkin, Suchir Balaji, Valerie Balcom,
  Paul Baltescu, Haiming Bao, Mohammad Bavarian, Jeff Belgum, Irwan Bello, Jake
  Berdine, Gabriel Bernadett-Shapiro, Christopher Berner, Lenny Bogdonoff, Oleg
  Boiko, Madelaine Boyd, Anna-Luisa Brakman, Greg Brockman, Tim Brooks, Miles
  Brundage, Kevin Button, Trevor Cai, Rosie Campbell, Andrew Cann, Brittany
  Carey, Chelsea Carlson, Rory Carmichael, Brooke Chan, Che Chang, Fotis
  Chantzis, Derek Chen, Sully Chen, Ruby Chen, Jason Chen, Mark Chen, Ben
  Chess, Chester Cho, Casey Chu, Hyung~Won Chung, Dave Cummings, Jeremiah
  Currier, Yunxing Dai, Cory Decareaux, Thomas Degry, Noah Deutsch, Damien
  Deville, Arka Dhar, David Dohan, Steve Dowling, Sheila Dunning, Adrien
  Ecoffet, Atty Eleti, Tyna Eloundou, David Farhi, Liam Fedus, Niko Felix,
  Sim{\'o}n~Posada Fishman, Juston Forte, Isabella Fulford, Leo Gao, Elie
  Georges, Christian Gibson, Vik Goel, Tarun Gogineni, Gabriel Goh, Rapha
  Gontijo-Lopes, Jonathan Gordon, Morgan Grafstein, Scott Gray, Ryan Greene,
  Joshua Gross, Shixiang~Shane Gu, Yufei Guo, Chris Hallacy, Jesse Han, Jeff
  Harris, Yuchen He, Mike Heaton, Johannes Heidecke, Chris Hesse, Alan Hickey,
  Wade Hickey, Peter Hoeschele, Brandon Houghton, Kenny Hsu, Shengli Hu, Xin
  Hu, Joost Huizinga, Shantanu Jain, Shawn Jain, Joanne Jang, Angela Jiang,
  Roger Jiang, Haozhun Jin, Denny Jin, Shino Jomoto, Billie Jonn, Heewoo Jun,
  Tomer Kaftan, {\L}ukasz Kaiser, Ali Kamali, Ingmar Kanitscheider,
  Nitish~Shirish Keskar, Tabarak Khan, Logan Kilpatrick, Jong~Wook Kim,
  Christina Kim, Yongjik Kim, Jan~Hendrik Kirchner, Jamie Kiros, Matt Knight,
  Daniel Kokotajlo, {\L}ukasz Kondraciuk, Andrew Kondrich, Aris Konstantinidis,
  Kyle Kosic, Gretchen Krueger, Vishal Kuo, Michael Lampe, Ikai Lan, Teddy Lee,
  Jan Leike, Jade Leung, Daniel Levy, Chak~Ming Li, Rachel Lim, Molly Lin,
  Stephanie Lin, Mateusz Litwin, Theresa Lopez, Ryan Lowe, Patricia Lue, Anna
  Makanju, Kim Malfacini, Sam Manning, Todor Markov, Yaniv Markovski, Bianca
  Martin, Katie Mayer, Andrew Mayne, Bob McGrew, Scott~Mayer McKinney,
  Christine McLeavey, Paul McMillan, Jake McNeil, David Medina, Aalok Mehta,
  Jacob Menick, Luke Metz, Andrey Mishchenko, Pamela Mishkin, Vinnie Monaco,
  Evan Morikawa, Daniel Mossing, Tong Mu, Mira Murati, Oleg Murk, David
  M{\'e}ly, Ashvin Nair, Reiichiro Nakano, Rajeev Nayak, Arvind Neelakantan,
  Richard Ngo, Hyeonwoo Noh, Long Ouyang, Cullen O'Keefe, Jakub Pachocki, Alex
  Paino, Joe Palermo, Ashley Pantuliano, Giambattista Parascandolo, Joel
  Parish, Emy Parparita, Alex Passos, Mikhail Pavlov, Andrew Peng, Adam
  Perelman, Filipe de~Avila Belbute~Peres, Michael Petrov, Henrique~Ponde
  de~Oliveira~Pinto, Michael, Pokorny, Michelle Pokrass, Vitchyr~H. Pong, Tolly
  Powell, Alethea Power, Boris Power, Elizabeth Proehl, Raul Puri, Alec
  Radford, Jack Rae, Aditya Ramesh, Cameron Raymond, Francis Real, Kendra
  Rimbach, Carl Ross, Bob Rotsted, Henri Roussez, Nick Ryder, Mario Saltarelli,
  Ted Sanders, Shibani Santurkar, Girish Sastry, Heather Schmidt, David
  Schnurr, John Schulman, Daniel Selsam, Kyla Sheppard, Toki Sherbakov, Jessica
  Shieh, Sarah Shoker, Pranav Shyam, Szymon Sidor, Eric Sigler, Maddie Simens,
  Jordan Sitkin, Katarina Slama, Ian Sohl, Benjamin Sokolowsky, Yang Song,
  Natalie Staudacher, Felipe~Petroski Such, Natalie Summers, Ilya Sutskever,
  Jie Tang, Nikolas Tezak, Madeleine~B. Thompson, Phil Tillet, Amin
  Tootoonchian, Elizabeth Tseng, Preston Tuggle, Nick Turley, Jerry Tworek,
  Juan Felipe~Cer{\'o}n Uribe, Andrea Vallone, Arun Vijayvergiya, Chelsea Voss,
  Carroll Wainwright, Justin~Jay Wang, Alvin Wang, Ben Wang, Jonathan Ward,
  Jason Wei, CJ~Weinmann, Akila Welihinda, Peter Welinder, Jiayi Weng, Lilian
  Weng, Matt Wiethoff, Dave Willner, Clemens Winter, Samuel Wolrich, Hannah
  Wong, Lauren Workman, Sherwin Wu, Jeff Wu, Michael Wu, Kai Xiao, Tao Xu,
  Sarah Yoo, Kevin Yu, Qiming Yuan, Wojciech Zaremba, Rowan Zellers, Chong
  Zhang, Marvin Zhang, Shengjia Zhao, Tianhao Zheng, Juntang Zhuang, William
  Zhuk, and Barret Zoph.
\newblock Gpt-4 technical report, 2024.

\bibitem[Radford et~al.(2019)Radford, Wu, Child, Luan, Amodei, Sutskever,
  et~al.]{radford2019language}
Alec Radford, Jeffrey Wu, Rewon Child, David Luan, Dario Amodei, Ilya
  Sutskever, et~al.
\newblock Language models are unsupervised multitask learners.
\newblock \emph{OpenAI blog}, 1\penalty0 (8):\penalty0 9, 2019.

\bibitem[Shuster et~al.(2021)Shuster, Poff, Chen, Kiela, and
  Weston]{shuster-etal-2021-retrieval-augmentation}
Kurt Shuster, Spencer Poff, Moya Chen, Douwe Kiela, and Jason Weston.
\newblock Retrieval augmentation reduces hallucination in conversation.
\newblock In Marie-Francine Moens, Xuanjing Huang, Lucia Specia, and Scott
  Wen-tau Yih, editors, \emph{Findings of the Association for Computational
  Linguistics: EMNLP 2021}, pages 3784--3803, Punta Cana, Dominican Republic,
  November 2021. Association for Computational Linguistics.
\newblock \doi{10.18653/v1/2021.findings-emnlp.320}.
\newblock URL \url{https://aclanthology.org/2021.findings-emnlp.320}.

\bibitem[Su et~al.(2023)Su, Lu, Pan, Murtadha, Wen, and Liu]{su2023roformer}
Jianlin Su, Yu~Lu, Shengfeng Pan, Ahmed Murtadha, Bo~Wen, and Yunfeng Liu.
\newblock Roformer: Enhanced transformer with rotary position embedding, 2023.

\bibitem[Su et~al.(2024)Su, Wang, Ai, HU, Wu, Zhou, and
  Liu]{su2024unsupervised}
Weihang Su, Changyue Wang, Qingyao Ai, Yiran HU, Zhijing Wu, Yujia Zhou, and
  Yiqun Liu.
\newblock Unsupervised real-time hallucination detection based on the internal
  states of large language models, 2024.

\bibitem[Thakur et~al.(2024)Thakur, Bonifacio, Zhang, Ogundepo, Kamalloo,
  Alfonso-Hermelo, Li, Liu, Chen, Rezagholizadeh, and Lin]{thakur2024nomiracl}
Nandan Thakur, Luiz Bonifacio, Xinyu Zhang, Odunayo Ogundepo, Ehsan Kamalloo,
  David Alfonso-Hermelo, Xiaoguang Li, Qun Liu, Boxing Chen, Mehdi
  Rezagholizadeh, and Jimmy Lin.
\newblock Nomiracl: Knowing when you don't know for robust multilingual
  retrieval-augmented generation, 2024.

\bibitem[Touvron et~al.(2023)Touvron, Martin, Stone, Albert, Almahairi, Babaei,
  Bashlykov, Batra, Bhargava, Bhosale, Bikel, Blecher, Ferrer, Chen, Cucurull,
  Esiobu, Fernandes, Fu, Fu, Fuller, Gao, Goswami, Goyal, Hartshorn, Hosseini,
  Hou, Inan, Kardas, Kerkez, Khabsa, Kloumann, Korenev, Koura, Lachaux, Lavril,
  Lee, Liskovich, Lu, Mao, Martinet, Mihaylov, Mishra, Molybog, Nie, Poulton,
  Reizenstein, Rungta, Saladi, Schelten, Silva, Smith, Subramanian, Tan, Tang,
  Taylor, Williams, Kuan, Xu, Yan, Zarov, Zhang, Fan, Kambadur, Narang,
  Rodriguez, Stojnic, Edunov, and Scialom]{touvron2023llama}
Hugo Touvron, Louis Martin, Kevin Stone, Peter Albert, Amjad Almahairi, Yasmine
  Babaei, Nikolay Bashlykov, Soumya Batra, Prajjwal Bhargava, Shruti Bhosale,
  Dan Bikel, Lukas Blecher, Cristian~Canton Ferrer, Moya Chen, Guillem
  Cucurull, David Esiobu, Jude Fernandes, Jeremy Fu, Wenyin Fu, Brian Fuller,
  Cynthia Gao, Vedanuj Goswami, Naman Goyal, Anthony Hartshorn, Saghar
  Hosseini, Rui Hou, Hakan Inan, Marcin Kardas, Viktor Kerkez, Madian Khabsa,
  Isabel Kloumann, Artem Korenev, Punit~Singh Koura, Marie-Anne Lachaux,
  Thibaut Lavril, Jenya Lee, Diana Liskovich, Yinghai Lu, Yuning Mao, Xavier
  Martinet, Todor Mihaylov, Pushkar Mishra, Igor Molybog, Yixin Nie, Andrew
  Poulton, Jeremy Reizenstein, Rashi Rungta, Kalyan Saladi, Alan Schelten, Ruan
  Silva, Eric~Michael Smith, Ranjan Subramanian, Xiaoqing~Ellen Tan, Binh Tang,
  Ross Taylor, Adina Williams, Jian~Xiang Kuan, Puxin Xu, Zheng Yan, Iliyan
  Zarov, Yuchen Zhang, Angela Fan, Melanie Kambadur, Sharan Narang, Aurelien
  Rodriguez, Robert Stojnic, Sergey Edunov, and Thomas Scialom.
\newblock Llama 2: Open foundation and fine-tuned chat models, 2023.

\bibitem[Wei et~al.(2023)Wei, Wang, Schuurmans, Bosma, Ichter, Xia, Chi, Le,
  and Zhou]{wei2023chainofthought}
Jason Wei, Xuezhi Wang, Dale Schuurmans, Maarten Bosma, Brian Ichter, Fei Xia,
  Ed~Chi, Quoc Le, and Denny Zhou.
\newblock Chain-of-thought prompting elicits reasoning in large language
  models, 2023.

\bibitem[Wu et~al.(2023)Wu, Zhu, Xu, Shum, Niu, Zhong, Song, and
  Zhang]{wu2023ragtruth}
Yuanhao Wu, Juno Zhu, Siliang Xu, Kashun Shum, Cheng Niu, Randy Zhong, Juntong
  Song, and Tong Zhang.
\newblock Ragtruth: A hallucination corpus for developing trustworthy
  retrieval-augmented language models, 2023.

\bibitem[Xu et~al.(2024)Xu, Cruz, Guevara, Wang, Deshpande, Wang, and
  Li]{xu2024rag}
Zhentao Xu, Mark~Jerome Cruz, Matthew Guevara, Tie Wang, Manasi Deshpande,
  Xiaofeng Wang, and Zheng Li.
\newblock Retrieval-augmented generation with knowledge graphs for customer
  service question answering.
\newblock In \emph{Proceedings of the 47th International ACM SIGIR Conference
  on Research and Development in Information Retrieval}, SIGIR '24, pages
  2905--2909, New York, NY, USA, 2024. Association for Computing Machinery.
\newblock ISBN 9798400704314.
\newblock \doi{10.1145/3626772.3661370}.
\newblock URL \url{https://doi.org/10.1145/3626772.3661370}.

\bibitem[Zhang and Sennrich(2019)]{zhang2019root}
Biao Zhang and Rico Sennrich.
\newblock Root mean square layer normalization, 2019.

\bibitem[Zhang et~al.(2023)Zhang, Li, Cui, Cai, Liu, Fu, Huang, Zhao, Zhang,
  Chen, Wang, Luu, Bi, Shi, and Shi]{zhang2023sirens}
Yue Zhang, Yafu Li, Leyang Cui, Deng Cai, Lemao Liu, Tingchen Fu, Xinting
  Huang, Enbo Zhao, Yu~Zhang, Yulong Chen, Longyue Wang, Anh~Tuan Luu, Wei Bi,
  Freda Shi, and Shuming Shi.
\newblock Siren's song in the ai ocean: A survey on hallucination in large
  language models, 2023.

\end{thebibliography}

\clearpage

\appendix
\section*{\centering \Huge \normalfont Appendix} 
\section{Prompt Template for Q\&A Generation}

\begin{listing}%
\caption{Prompt template for generating a question and the corresponding quoted answer based on a passage and its previous context from a Wikipedia article. The three placeholders `title`, `section\_before\_passage`, and `passage\_text` are substituted with the particular content of a passage and its article title. The system message is embraced by \texttt{<<sys>>} tags.}%
\label{lst:prompt_template_wiki2qna}%
\begin{lstlisting}
<<sys>>You are perfect at creating a question based on a sentence and its previous context. You also cite the answer to those questions from the given sentence.<</sys>>
### PREVIOUS CONTEXT
Title: '{title}'
{section_before_passage}

### SENTENCE
{passage_text}

### OBJECTIVE
Write a question solely based on the given SENTENCE. This SENTENCE contains the definite answer which you also quote. This quote is definitely part of the SENTENCE. The question does not have the same wording as the SENTENCE. The question is 'globally' phrased and not 'locally', meaning that the question can be asked in a Retrieval Augmented Generation application.

### RESPONSE
The json format of your response should look like this:
```json
{
    "answer_quote": <answer copied from the sentence (as brief as possible)>,
    "question": <question that is answer with the answer_quote>
}
```
Ensure your response can be parsed using Python json.loads
\end{lstlisting}
\end{listing}

\clearpage

\section{Prompt Templates Used in HalluRAG}

\begin{listing}%
\caption{Adjusted prompt template from the langchain hub (id: "rlm/rag-prompt"; retrieved on May 19, 2024). The wording has been changed from "three sentences at maximum" to "as few sentences as possible" to avoid forcing an LLM to generate a fixed amount of sentences. The two placeholders `question` and `context` are replaced with the real corresponding text.}%
\label{lst:prompt_template_langchain_hub}%
\begin{lstlisting}
You are an assistant for question-answering tasks. Use the following pieces of retrieved context to answer the question. If you don't know the answer, just say that you don't know. Use as few sentences as possible and keep the answer concise.
Question: {question}
Context: {context}
Answer:
\end{lstlisting}
\end{listing}

\begin{listing}%
\caption{Prompt template 1 with an additional system message embraced by two \texttt{<<sys>>} tags. The two placeholders `question` and `context` are replaced with the real corresponding text.}%
\label{lst:prompt_template_1}%
\begin{lstlisting}
<<sys>>You are a helpful, respectful, and honest assistant for a question-answering task. You are provided pieces of context that MIGHT contain the answer to the question. Your concise answer should solely be based on these pieces. Always answer as helpfully as possible, while being safe. Your answer should not include any harmful, unethical, racist, sexist, toxic, dangerous, or illegal content. If a question does not make any sense, or is not factually coherent, explain why instead of answering something not correct. If you don't know the answer to a question, state that so you don't share false information. Do not refer to chunks literally. Do not use the word 'chunk', just use their information for your answer. Do NOT start with 'Based on...'<</sys>>
Your knowledge is limited to only this information:
{context}
QUESTION: {question}
BRIEF ANSWER:
\end{lstlisting}
\end{listing}

\begin{listing}%
\caption{Prompt template 2 with an additional system message embraced by two \texttt{<<sys>>} tags and a more detailed instruction on how to answer properly. The two placeholders `question` and `context` are replaced with the real corresponding text.}%
\label{lst:prompt_template_2}%
\begin{lstlisting}
<<sys>>You are a helpful, respectful, and honest assistant for a question-answering task. You are provided pieces of context that MIGHT contain the answer to the question. Your concise answer should solely be based on these pieces. Always answer as helpfully as possible, while being safe. Your answer should not include any harmful, unethical, racist, sexist, toxic, dangerous, or illegal content. If a question does not make any sense, or is not factually coherent, explain why instead of answering something not correct. If you don't know the answer to a question, state that so you don't share false information.<</sys>>
Only use the information included in these chunks to answer the question:
{context}
QUESTION: {question}
REMINDER: If no chunk contains the information asked for, briefly explain that you cannot answer the question. However, do not refer to chunks literally. Do not use the word 'chunk' or that chunks were provided to you, just use their information to answer the QUESTION. Do NOT start with 'Based on...'
BRIEF RESPONSE:
\end{lstlisting}
\end{listing}

\clearpage
\section{Auto-Labeling Procedure and Results}

\begin{table}[h]
\centering
\begin{tabular}{|l|r|r|r|r||r|}
\hline
\textbf{Answerability} & \textbf{C} & \textbf{G} & \textbf{F} & \textbf{IDK} & \textbf{Hallucinated} \\
\hline
answerable  & 1 & 0 & * & * & 1 \\
            & 1 & 1 & 0 & * & 1 \\
            & 1 & 1 & 1 & 0 & 1 \\
            & 1 & 1 & 1 & 1 & None \\
            & 0 & 1 & 1 & 1 & None \\
            & 0 & 1 & 1 & 0 & 0 \\
            & 0 & 1 & 0 & 1 & 1 \\
            & 0 & 0 & 1 & 1 & 1 \\
            & 0 & 0 & 0 & 1 & 1 \\
            & 0 & 0 & 1 & 0 & 1 \\
            & 0 & * & 0 & 0 & 0 \\
\hline
unanswerable& 1 & 0 & * & * & 1 \\
            & 1 & 1 & 0 & * & 1 \\
            & 1 & 1 & 1 & 0 & 1 \\
            & 0 & 0 & 1 & * & 1 \\
            & 1 & 1 & 1 & 1 & None \\
            & 0 & 1 & 1 & 1 & None \\
            & 0 & 1 & 0 & 1 & 0 \\
            & 0 & 0 & 0 & 1 & 0 \\
            & 0 & 1 & 1 & 0 & None \\
            & 0 & * & 0 & 0 & 0 \\
\hline
\end{tabular}
\caption{Truth table for determining if a sentence is hallucinated (1) or non-hallucinated (0), given the answerability and the booleans \texttt{conflicting}~(C), \texttt{grounded}~(G), \texttt{has\_factual\_information}~(F), and \texttt{no\_clear\_answer}~(IDK). 'None' means that the case should not be possible to achieve. These cases are ignored in training, validation, and testing.}
\label{tab:hallu_mapping}
\end{table}

\begin{table}[h]
\centering
\begin{tabular}{|l||r|r|r|}
    \hline
    \textbf{Model} & \textbf{Sentences} & \textbf{Valid} & \textbf{Valid (\%)} \\
    \hline
    L2-7B & 2201 & 1932 & 87.78 \\
    L2-7B (float8) & 2223 & 1989 & 89.47 \\
    L2-7B (int4) & 3577 & 3282 & 91.75 \\
    L2-7B (int8) & 2176 & 1930 & 88.69 \\
    \hline
    
    L2-13B (float8) & 2188 & 1963 & 89.72 \\
    L2-13B (int4) & 2153 & 1941 & 90.15 \\
    L2-13B (int8) & 2199 & 1991 & 90.54 \\
    \hline
    
    M-7B & 1193 & 1152 & 96.56 \\
    M-7B (float8) & 1202 & 1158 & 96.34 \\
    M-7B (int4) & 1232 & 1205 & 97.81 \\
    M-7B (int8) & 1212 & 1188 & 98.02 \\
    \hline
\end{tabular}
\caption{Results of GPT-4o labeling sentences from various LLM configurations, showing the total and valid sentence counts. Abbreviations: LLaMA-2-7B-Chat-HF as L2-7B, LLaMA-2-13B-Chat-HF as L2-13B, Mistral-7B-Instruct-v0.1 as M-7B.}
\label{tab:auto_labeling_results}
\end{table}

\begin{table}[h]
\centering
\begin{tabular}{|l||r|r|r|r|}
    \hline
    \textbf{Model} & \textbf{All (\%)} & \textbf{hub (\%)} & \textbf{1 (\%)} & \textbf{2 (\%)} \\
    \hline
    L2-7B & 21.07 & 30.92 & 17.76 & 16.35 \\
    L2-7B (float8) & 21.57 & 32.46 & 17.53 & 16.72 \\
    L2-7B (int4) & 21.18 & 40.92 & 16.72 & 13.27 \\
    L2-7B (int8) & 21.24 & 32.00 & 18.39 & 15.22 \\
    \hline
    
    L2-13B (float8) & 15.49 & 21.29 & 9.49 & 15.25 \\
    L2-13B (int4) & 15.46 & 16.35 & 10.08 & 20.36 \\
    L2-13B (int8) & 15.82 & 23.70 & 9.77 & 13.21 \\
    \hline
    
    M-7B & 10.24 & 14.96 & 9.19 & 6.67 \\
    M-7B (float8) & 9.15 & 15.42 & 8.44 & 3.59 \\
    M-7B (int4) & 11.45 & 16.80 & 9.63 & 8.23 \\
    M-7B (int8) & 10.86 & 15.21 & 10.33 & 7.20 \\
    \hline
\end{tabular}
\caption{Hallucination rates (HRs) of various LLM configurations. Abbreviations: LLaMA-2-7B-Chat-HF as L2-7B, LLaMA-2-13B-Chat-HF as L2-13B, Mistral-7B-Instruct-v0.1 as M-7B. `All` represents the HR independent of the prompt template, which is also broken down by prompt template: `hub` for "rlm/rag-prompt" from Langchain hub (Listing~\ref{lst:prompt_template_langchain_hub}), `1` (Listing~\ref{lst:prompt_template_1}), and `2` (Listing~\ref{lst:prompt_template_2}).}
\label{tab:hallu_rates}
\end{table}

\clearpage

\section{Further Results on Accuracies of HalluRAG-Trained MLPs}

\begin{table}[h]
\centering
\begin{tabular}{|l||r|r|r|r|}
\hline
\textbf{Internal State} & \textbf{All (\%)} & \textbf{float8 (\%)} & \textbf{int8 (\%)} & \textbf{int4 (\%)} \\
\hline
\textbf{LLaMA-2-13B-Chat-HF} & & & &  \\
cev (middle) & 49.75±2.29 &  52.34±3.10 & 44.92±1.01 & 56.87±2.69 \\
cev (last) & 50.40±0.25 &  52.34±4.60 & 54.52±7.21 & 59.68±1.29 \\
iav (middle) & 50.48±0.00 &  54.71±1.31 & 45.48±2.20 & 50.78±1.07 \\
iav (last) & 50.48±0.00 &  52.78±0.57 & 46.03±0.54 & 52.52±3.16 \\
\hline
\end{tabular}
\caption{Average test accuracies and standard deviations of training ten MLPs on the internal states of the responses of LLaMA-2-13B-Chat-HF taken from HalluRAG. The column names indicate the quantization, where 'All' means all quantizations together (float8, int8, and int4). All internal states have been extracted from the last token's middle and last decoder block or 'layer.' We abbreviate contextualized embedding vectors as 'CEV' and intermediate activation values as 'IAV'.}
\label{tab:hallurag-l13}
\end{table}

\begin{table}[h]
\centering
\begin{tabular}{|l||r|r|r|r|r|}
\hline
\textbf{Internal State} & \textbf{All (\%)} & \textbf{None (\%)} & \textbf{float8 (\%)} & \textbf{int8 (\%)} & \textbf{int4 (\%)} \\
\hline

\textbf{LLaMA-2-7B-Chat-HF} & & & & & \\
cev (middle) \& cev (last) & 59.6±1.9 & 55.82±4.14 & 57.79±2.92 & 58.85±1.55 & 69.9±2.09 \\
cev (middle) \& iav (middle) & 65.3±0.99 & 58.33±0.8 & 63.62±1.47 & 62.04±1.97 & 71.21±0.54 \\
cev (middle) \& iav (last) & 61.88±1.87 & 58.07±1.55 & 64.61±4.0 & 60.17±1.36 & 73.82±0.85 \\
cev (last) \& iav (middle) & 59.81±1.63 & 57.89±1.8 & 58.11±2.64 & 59.13±1.19 & 70.04±1.48 \\
cev (last) \& iav (last) & 59.18±1.65 & 55.87±4.52 & 57.3±3.92 & 57.53±1.18 & 70.68±3.12 \\
iav (middle) \& iav (last) & 60.54±2.33 & 57.58±0.83 & 61.1±2.48 & 61.83±2.06 & 74.5±0.87 \\
\hline
all four & 61.4±1.41 & 54.96±2.92 & 58.14±2.37 & 58.4±1.45 & 72.01±1.59 \\
\hline

\textbf{LLaMA-2-13B-Chat-HF} & & & & & \\
cev (middle) \& cev (last) & 50.67±2.55 & - & 52.78±4.79 & 46.43±3.59 & 57.04±0.96 \\
cev (middle) \& iav (middle) & 50.99±0.62 & - & 51.41±3.34 & 46.73±2.91 & 57.42±1.45 \\
cev (middle) \& iav (last) & 49.12±1.33 & - & 51.8±3.16 & 48.87±4.1 & 58.09±2.15 \\
cev (last) \& iav (middle) & 50.48±0.0 & - & 50.77±0.69 & 48.15±1.26 & 54.12±4.54  \\
cev (last) \& iav (last) & 50.7±0.43 & - & 50.52±0.35 & 46.73±1.21 & 57.71±4.81 \\
iav (middle) \& iav (last) & 50.5±0.06 & - & 51.02±1.31 & 46.21±2.22 & 50.42±0.0  \\
\hline
all four & 49.42±1.74 & - & 53.8±4.32 & 47.72±4.74 & 58.09±2.4 \\
\hline

\textbf{Mistral-7B-Instruct-v0.1}  & & & & & \\
cev (middle) \& cev (last) & 74.02±2.75 & \textit{53.57±4.26} & \textit{62.4±4.86} & \textit{71.92±1.92} & \textit{73.81±5.71} \\
cev (middle) \& iav (middle) & 71.23±1.54 & \textit{54.68±7.77} & \textit{59.13±5.71} & \textit{67.14±0.28} & \textit{70.47±4.25} \\
cev (middle) \& iav (last) & 73.41±1.35 & \textit{60.78±0.38} & \textit{70.8±6.04} & \textit{68.54±2.51} & \textit{81.19±0.59} \\
cev (last) \& iav (middle) & 73.23±2.91 & \textit{55.78±7.38} & \textit{64.94±8.44} & \textit{74.19±8.05} & \textit{77.28±5.85} \\
cev (last) \& iav (last) & 72.26±2.85 & \textit{54.9±5.11} & \textit{66.86±6.82} & \textit{74.55±5.34} & \textit{76.09±3.78} \\
iav (middle) \& iav (last) & 73.64±3.39 & \textit{53.73±3.57} & \textit{66.63±8.11} & \textit{64.74±2.52} & \textit{81.06±3.86} \\
\hline
all four & 74.39±3.42 & \textit{52.21±3.51} & \textit{67.4±5.37} & \textit{68.02±8.36} & \textit{80.28±1.6} \\
\hline
\end{tabular}
\caption{Average test accuracies and standard deviations of training ten MLPs on combined internal states of responses taken from HalluRAG. The column names indicate the quantization, where 'None' means the use of no quantization and 'All' means all quantizations together (None, float8, int8, and int4). All internal states have been extracted from the last token's middle and last decoder block or 'layer.' We abbreviate contextualized embedding vectors as 'CEV' and intermediate activation values as 'IAV'. The values based on a single quantized Mistral configuration are italic, since their test sets turn out insufficiently big for being analyzed.}
\label{tab:hallurag-combinations}
\end{table}

\clearpage

\section{Accuracies of Answerability-Separated HalluRAG-Trained MLPs}

\begin{table}[h]
\centering
\begin{tabular}{|l||r|r|r|r|r|}
\hline
\textbf{Internal State} & \textbf{All (\%)} & \textbf{None (\%)} & \textbf{float8 (\%)} & \textbf{int8 (\%)} & \textbf{int4 (\%)} \\
\hline

\textbf{LLaMA-2-7B-Chat-HF} & & & & & \\
cev (middle) & 75.21±0.45 & 67.12±2.79 & 75.60±1.19 & 79.50±0.31 & 87.12±0.85 \\
cev (last) & 77.64±0.63 & 69.28±0.58 & 74.53±1.02 & 75.72±1.35 & 89.62±0.82 \\
iav (middle) & 75.52±0.59 & 69.93±0.00 & 73.77±0.40 & 78.74±0.97 & 84.96±0.45 \\
iav (last) & 78.22±0.54 & 69.48±0.51 & 75.09±0.42 & 77.48±2.46 & 90.61±1.28 \\
\hline
cev (middle) \& cev (last) & 78.18±0.44 & 68.89±0.43 & 74.53±1.81 & 77.23±0.79 & 89.77±0.2 \\
cev (middle) \& iav (middle) & 75.64±0.58 & 70.2±0.32 & 76.04±0.44 & 79.18±0.91 & 86.33±0.46 \\
cev (middle) \& iav (last) & 78.92±0.24 & 70.46±0.7 & 74.59±1.2 & 79.12±0.88 & 90.68±0.39 \\
cev (last) \& iav (middle) & 77.38±0.62 & 68.82±0.51 & 73.77±1.52 & 76.92±1.41 & 89.81±0.27  \\
cev (last) \& iav (last) & 78.28±0.37 & 69.02±0.32 & 73.52±1.67 & 75.72±0.31 & 89.55±0.49 \\
iav (middle) \& iav (last) & 78.04±0.8 & 69.93±0.51 & 74.78±1.77 & 77.55±1.44 & 90.87±0.43  \\
\hline
all four & 77.92±0.78 & 68.95±0.33 & 73.27±1.76 & 76.29±0.75 & 89.96±0.19 \\
\hline

\textbf{LLaMA-2-13B-Chat-HF}  & & & & & \\
cev (middle) & 49.68±6.81 & - & 76.67±10.56 & 67.05±9.74 & 47.64±8.35 \\
cev (last) & 45.27±1.61 & - & 71.60±13.39 & 71.41±4.58 & 47.91±4.86 \\
iav (middle) & 56.98±1.61 & - & 82.51±0.67 & 71.69±0.08 & 50.00±0.00 \\
iav (last) & 58.65±2.91 & - & 82.53±0.39 & 73.79±0.61 & 48.82±2.57 \\
\hline
cev (middle) \& cev (last) & 45.23±1.73 & - & 73.93±11.99 & 67.22±9.61 & 47.45±6.36 \\
cev (middle) \& iav (middle) & 44.43±0.59 & - & 76.3±8.97 & 67.1±6.91 & 47.32±4.46 \\
cev (middle) \& iav (last) & 48.91±7.57 & - & 80.44±3.07 & 66.57±9.34 & 45.95±4.92 \\
cev (last) \& iav (middle) & 46.89±1.58 & - & 82.35±2.58 & 71.21±1.13 & 50.0±0.0  \\
cev (last) \& iav (last) & 47.69±1.81 & - & 81.37±8.87 & 71.57±1.48 & 50.0±0.0 \\
iav (middle) \& iav (last) & 57.34±1.62 & - & 82.74±0.25 & 69.39±6.97 & 50.0±0.0  \\
\hline
all four & 45.24±1.67 & - & 77.8±10.06 & 68.74±6.77 & 43.55±5.77 \\
\hline

\textbf{Mistral-7B-Instruct-v0.1} & & & & & \\
cev (middle) & 69.56±0.16 & \textit{83.05±2.37} & \textit{70.23±0.28} & \textit{82.29±0.00} & \textit{85.63±11.27} \\
cev (last) & 70.66±0.53 & \textit{80.60±3.96} & \textit{68.52±1.42} & \textit{82.29±0.00} & \textit{88.06±11.38} \\
iav (middle) & 68.55±2.08 & \textit{83.80±0.11} & \textit{70.45±0.00} & \textit{82.29±0.00} & \textit{92.51±10.59} \\
iav (last) & 70.38±0.33 & \textit{79.10±3.95} & \textit{69.15±0.44} & \textit{82.29±0.00} & \textit{92.08±10.37} \\
\hline
cev (middle) \& cev (last) & 70.63±0.42 & \textit{80.64±3.99} & \textit{68.81±1.47} & \textit{82.29±0.0} & \textit{94.85±9.04} \\
cev (middle) \& iav (middle) & 70.12±0.3 & \textit{83.8±0.11} & \textit{70.45±0.0} & \textit{82.29±0.0} & \textit{97.13±6.93} \\
cev (middle) \& iav (last) & 70.37±0.43 & \textit{79.14±3.92} & \textit{69.09±0.28} & \textit{82.29±0.0} & \textit{98.87±0.0} \\
cev (last) \& iav (middle) & 70.55±0.54 & \textit{83.05±2.37} & \textit{69.83±0.59} & \textit{82.29±0.0} & \textit{97.44±6.94}  \\
cev (last) \& iav (last) & 70.69±0.38 & \textit{80.71±3.9} & \textit{69.32±0.36} & \textit{82.29±0.0} & \textit{99.24±0.22} \\
iav (middle) \& iav (last) & 70.67±0.43 & \textit{83.87±0.11} & \textit{69.77±0.56} & \textit{82.29±0.0} & \textit{99.44±0.22}  \\
\hline
all four & 70.22±0.41 & \textit{81.5±3.64} & \textit{69.72±0.44} & \textit{82.51±0.66} & \textit{97.44±6.94} \\
\hline

\end{tabular}
\caption{Average test accuracies and standard deviations of training ten MLPs on singular and combined internal states of the responses based on all \textbf{answerable} questions of HalluRAG. The column names indicate the quantization, where 'None' means the usage of no quantization and 'All' means all quantizations together (None, float8, int8, and int4). All internal states have been extracted from the last token's middle and last decoder block or 'layer.' We abbreviate contextualized embedding vectors as 'CEV' and intermediate activation values as 'IAV'. The values based on a single quantized Mistral configuration are italic, since their test sets turn out insufficiently big for being analyzed.}
\label{tab:hallurag-answerable-all}
\end{table}

\begin{table}[h]
\centering
\begin{tabular}{|l||r|r|r|r|r|}
\hline
\textbf{Internal State} & \textbf{All (\%)} & \textbf{None (\%)} & \textbf{float8 (\%)} & \textbf{int8 (\%)} & \textbf{int4 (\%)} \\
\hline

\textbf{LLaMA-2-7B-Chat-HF} & & & & & \\ 
cev (middle) & 82.89±0.43  & 80.46±0.25  & 85.33±1.12  &  91.91±0.50  &  94.34±1.76 \\
cev (last) & 87.13±0.40  &  84.09±0.13  &  89.87±1.71  &  93.95±0.47  &  94.99±3.00 \\
iav (middle) & 81.98±0.83  & 80.63±0.61  & 88.77±0.58  & 92.17±0.25  & 90.53±2.59 \\
iav (last) & 86.26±0.14  & 83.56±0.41  & 89.97±1.99  & 92.68±3.53  & 90.53±5.75 \\
\hline
cev (middle) \& cev (last) & 87.98±0.67 & 83.99±0.3 & 90.03±2.15 & 93.63±0.4 & 96.07±0.09 \\
cev (middle) \& iav (middle) & 82.85±0.27 & 80.5±0.27 & 89.23±0.21 & 92.2±0.33 & 94.11±1.14 \\
cev (middle) \& iav (last) & 87.26±0.11 & 83.63±0.22 & 90.1±1.54 & 94.14±0.29 & 95.53±0.26 \\
cev (last) \& iav (middle) & 87.59±0.22 & 84.09±0.2 & 90.6±2.48 & 93.28±0.54 & 94.31±3.47 \\
cev (last) \& iav (last) & 87.22±0.41 & 84.16±0.0 & 91.13±2.31 & 93.69±0.66 & 93.94±4.04 \\
iav (middle) \& iav (last) & 87.11±0.05 & 83.56±1.03 & 92.57±1.64 & 94.27±0.0 & 95.73±0.13 \\
\hline
all four & 87.56±0.22 & 84.13±0.23 & 90.33±2.21 & 93.89 ±0.71 & 94.89 ±3.0 \\
\hline

\textbf{LLaMA-2-13B-Chat-HF}  & & & & & \\
cev (middle) & 47.56±4.79 & - &  51.47±3.27 & 49.13±6.66 & 48.78±5.70 \\
cev (last) & 50.53±0.27 & - &  50.20±0.00 & 52.77±5.50 & 49.28±5.17 \\
iav (middle) & 50.44±0.00 & - &  50.20±0.00 & 50.38±0.00 & 50.36±0.00 \\
iav (last) & 50.44±0.00 & - &  50.20±0.00 & 50.38±0.00 & 50.36±0.00 \\
\hline
cev (middle) \& cev (last) & 48.71±4.3 & - & 50.56±2.61 & 53.33±5.21 & 50.83±4.2 \\
cev (middle) \& iav (middle) & 50.08±3.76 & - & 52.67±2.89 & 53.18±6.02 & 51.55±5.19 \\
cev (middle) \& iav (last) & 45.7±3.16 & - & 50.88±3.29 & 55.61±5.54 & 51.19±5.09 \\
cev (last) \& iav (middle) & 50.44±0.0 & - & 49.56±2.52 & 50.95±1.7 & 50.36±0.0  \\
cev (last) \& iav (last) & 50.44±0.0 & - & 49.52±2.03 & 50.38±0.0 & 50.36±0.0 \\
iav (middle) \& iav (last) & 50.75±0.91 & - & 50.2±0.0 & 50.38±0.0 & 50.36±0.0  \\
\hline
all four & 46.75±4.47 & - & 50.44±2.21 & 50.57±5.31 & 50.47±4.49 \\
\hline

\textbf{Mistral-7B-Instruct-v0.1} & & & & & \\
cev (middle) & 99.55±0.24 & \textit{99.67±0.00} & \textit{99.33±0.00} & \textit{99.38±0.10} & \textit{98.15±0.00} \\
cev (last) & 100.00±0.00 & \textit{99.97±0.10} & \textit{100.00±0.00} & \textit{100.00±0.00} & \textit{99.57±0.28} \\
iav (middle) & 99.87±0.04 & \textit{99.50±0.17} & \textit{98.73±2.05} & \textit{98.43±1.84} & \textit{98.52±0.30} \\
iav (last) & 100.00±0.00 & \textit{99.97±0.10} & \textit{100.00±0.00} & \textit{99.90±0.15} & \textit{99.63±0.30} \\
\hline
cev (middle) \& cev (last) & 100.0±0.0 & \textit{100.0±0.0} & \textit{100.0±0.0} & \textit{100.0±0.0} & \textit{99.75±0.3} \\
cev (middle) \& iav (middle) & 99.82±0.09 & \textit{99.43±0.15} & \textit{99.53±0.31} & \textit{99.41±0.13} & \textit{98.33±0.28} \\
cev (middle) \& iav (last) & 100.0±0.0 & \textit{99.83±0.17} & \textit{100.0±0.0} & \textit{99.77±0.21} & \textit{99.69±0.31} \\
cev (last) \& iav (middle) & 100.0±0.0 & \textit{99.9±0.15} & \textit{100.0±0.0} & \textit{99.93±0.13} & \textit{99.63±0.3} \\
cev (last) \& iav (last) & 100.0±0.0 & \textit{100.0±0.0} & \textit{100.0±0.0} & \textit{99.87±0.26} & \textit{99.69±0.31} \\
iav (middle) \& iav (last) & 100.0±0.0 & \textit{99.7±0.1} & \textit{100.0±0.0} & \textit{99.61±0.29} & \textit{99.63±0.3} \\
\hline
all four & 100.0±0.0 & \textit{99.8±0.22} & \textit{100.0±0.0} & \textit{99.93±0.13} & \textit{99.69±0.31} \\
\hline
\end{tabular}
\caption{Average test accuracies and standard deviations of training ten MLPs on singular and combined internal states of the responses based on all \textbf{unanswerable} questions of HalluRAG. The column names indicate the quantization, where 'None' means the usage of no quantization and 'All' means all quantizations together (None, float8, int8, and int4). All internal states have been extracted from the last token's middle and last decoder block or 'layer.' We abbreviate contextualized embedding vectors as 'CEV' and intermediate activation values as 'IAV'. The values based on a single quantized Mistral configuration are italic, since their test sets turn out insufficiently big for being analyzed.}
\label{tab:hallurag-unanswerable-all}
\end{table}

\clearpage

\section{Results for Generalization between Datasets}

\begin{table}[h]
\centering
\begin{tabular}{|l||r|r|r|}
\hline
\textbf{Parameter} & \textbf{Withheld} & \textbf{Test Acc. (\%)} & \textbf{Diff. (\%)} \\
\hline
\textbf{answerable} & true & 64.64±0.54 & 0.77 \\
    & false & 48.41±0.64 & 17.00 \\
\hline
\textbf{chunk} & 350 & 64.17±0.22 & 1.24 \\
\textbf{size} & 550 & 63.04±0.53 & 2.37 \\
    & 750 & 65.24±0.58 & 0.17 \\
\hline
\textbf{chunks} & 1 & 66.73±1.03 & -1.32 \\
\textbf{per} & 3 & 63.64±0.41 & 1.77 \\
\textbf{prompt} & 5 & 65.37±0.62 & 0.04 \\
\hline
\textbf{prompt} & hub & 65.80±0.79 & -0.39 \\
\textbf{template} & 1 & 65.36±0.59 & 0.05 \\
    & 2 & 64.19±0.57 & 1.22 \\
\hline
\end{tabular}
\caption{Average test accuracies and standard deviations for different parameters that have been left out in the training of ten MLPs on HalluRAG. The internal states used are the contextualized embedding vectors of the last token's middle layer (cev (middle)) of all LLaMA-2-7B-Chat-HF models (None, float8, int8, and int4). The column 'Diff.' contains the difference between the reference accuracy (65.41\%) and the corresponding accuracy in the row.}
\label{tab:test_accuracies_left_out}
\end{table}

\begin{table}[h]
\centering
\begin{tabular}{|l||r|r|}
\hline
\textbf{Internal State} & \textbf{H-H-R (\%)} & \textbf{R-R-H (\%)} \\
\hline
\textbf{LLaMA-2-13B-Chat-HF} & & \\
cev (middle) & 50.01±1.05 & 45.75±1.38 \\
cev (last) & 50.00±0.00  & 43.02±1.08 \\
iav (middle) & 50.00±0.00  & 47.17±1.22 \\
iav (last) & 50.00±0.00  & 41.05±0.48 \\
\hline
\end{tabular}
\caption{Testing on the opposite dataset (LLaMA-2-13B-Chat-HF): Average test accuracies and standard deviations of testing ten MLPs, which on the one side have been trained and validated on HalluRAG but tested on RAGTruth (H-H-R), and on the other side trained and validated on RAGTruth but tested on HalluRAG (R-R-H). All quantizations together were used. All internal states have been extracted from the last token's middle and last decoder block or 'layer.' We abbreviate contextualized embedding vectors as 'CEV' and intermediate activation values as 'IAV'.}
\label{tab:tested-opposite-l13}
\end{table}

\end{document}